\documentclass[11pt,a4paper]{article}
\usepackage[hyperref]{acl2020}
\usepackage{times}
\usepackage{latexsym}

\usepackage{microtype}

\aclfinalcopy 

\usepackage{times}
\usepackage{soul}
\usepackage{url}
\usepackage{graphicx}
\usepackage{amsmath}
\usepackage{booktabs}
\usepackage{algorithm}
\usepackage{algorithmic}
\usepackage{times}
\usepackage{xcolor}

\usepackage{color,url}
\usepackage{times}
\usepackage{helvet}
\usepackage{courier}
\usepackage{amsfonts}
\usepackage{amsmath}
\usepackage{amssymb}
\usepackage{graphicx}
\usepackage{graphics}
\usepackage{epsfig}
\usepackage{mathptmx}
\usepackage{url}
\usepackage[l]{floatflt}
\usepackage{rotating}
\usepackage{subfigure}
\usepackage{multicol}
\usepackage{algorithm}
\usepackage{algorithmic}
\usepackage{verbatim}
\usepackage{listings}
\usepackage{chngpage}
\usepackage{wrapfig,lipsum}
\usepackage{booktabs}
\usepackage{threeparttable}
\usepackage{multirow}
\usepackage{flushend}

\usepackage{microtype}

\DeclareMathAlphabet{\mathcal}{OMS}{cmsy}{m}{n}

\floatname{algorithm}{Procedure}

\long\def\c#1{{\footnotesize{\fontfamily{pcr}\selectfont{#1}}}}

\title{Learning and Reasoning for Robot Dialog and Navigation Tasks}

\author{Keting Lu\textsuperscript{1} \quad Shiqi Zhang\textsuperscript{2} \quad Peter Stone\textsuperscript{3} \quad Xiaoping Chen\textsuperscript{1} \\
  \textsuperscript{1} USTC \quad
  \textsuperscript{2} SUNY Binghamton \quad
  \textsuperscript{3} UT Austin\\
  \texttt{ktlu@mail.ustc.edu.cn}; 
  \texttt{zhangs@binghamton.edu}; \\
  \texttt{pstone@cs.utexas.edu}; 
  \texttt{xpchen@ustc.edu.cn}
}

\date{}

\begin{document}

\maketitle

\begin{abstract}
    Reinforcement learning and probabilistic reasoning algorithms aim at learning from interaction experiences and reasoning with probabilistic contextual knowledge respectively. 
    In this research, we develop algorithms for robot task completions, while looking into the complementary strengths of reinforcement learning and probabilistic reasoning techniques. 
    The robots learn from trial-and-error experiences to augment their declarative knowledge base, and the augmented knowledge can be used for speeding up the learning process in potentially different tasks. 
    We have implemented and evaluated the developed algorithms using mobile robots conducting dialog and navigation tasks. 
    From the results, we see that our robot's performance can be improved by both reasoning with human knowledge and learning from task-completion experience. 
    More interestingly, the robot was able to learn from navigation tasks to improve its dialog strategies. 
\end{abstract}

\section{Introduction}

Knowledge representation and reasoning (KRR) and reinforcement learning (RL) are two important research areas in artificial intelligence (AI) and have been applied to a variety of problems in robotics. 
On the one hand, KRR research aims to concisely represent knowledge, and robustly draw conclusions with the knowledge (or generate new knowledge). 
Knowledge in KRR is typically provided by human experts in the form of declarative rules. 
Although KRR paradigms are strong in representing and reasoning with knowledge in a variety of forms, they are not designed for (and hence not good at) learning from experiences of accomplishing the tasks. 
On the other hand, RL algorithms enable agents to learn by interacting with an environment, and RL agents are good at learning action policies from trial-and-error experiences toward maximizing long-term rewards under uncertainty, but they are ill-equipped to utilize declarative knowledge from human experts. 
Motivated by the complementary features of KRR and RL, we aim at a framework that integrates both paradigms to enable agents (robots in our case) to simultaneously reason with declarative knowledge and learn by interacting with an environment. 

Most KRR paradigms support the representation and reasoning of knowledge in logical form, e.g., Prolog-style. 
More recently, researchers have developed hybrid KRR paradigms that support both logical and probabilistic knowledge~\cite{richardson2006markov,bach2017hinge,wang2019bridging}. 
Such logical-probabilistic KRR paradigms can be used for a variety of reasoning tasks. 
We use P-log~\cite{baral2009probabilistic} in this work to represent and reason with both human knowledge and the knowledge from RL. 
The reasoning results are then used by our robot to compute action policies at runtime. 

Reinforcement learning (RL) algorithms can be used to help robots learn action policies from the experience of interacting with the real world~\cite{sutton2018reinforcement}.
We use model-based RL in this work, because the learned world model can be used to update the robot's declarative knowledge base and combined with human knowledge.

\paragraph{Theoretical Contribution:} In this paper, we develop a learning and reasoning framework (called KRR-RL) that integrates logical-probabilistic KRR and model-based RL. 
The KRR component reasons with the qualitative knowledge from humans (e.g., it is difficult for a robot to navigate through a busy area) and the quantitative knowledge from model-based RL (e.g., a navigation action's success rate in the form of a probability). 
The hybrid knowledge is then used for computing action policies at runtime by planning with task-oriented partial world models. 
KRR-RL enables a robot to: 
i) represent the probabilistic knowledge (i.e., world dynamics) learned from RL in declarative form; 
ii) unify and reason with both human knowledge and the knowledge from RL; and
iii) compute policies at runtime by dynamically constructing task-oriented partial world models. 

\paragraph{Application Domain:}
We use a robot delivery domain for demonstration and evaluation purposes, where the robot needs to \textbf{dialog with people} to figure out the delivery task's goal location, and then physically \textbf{take navigation actions} to complete the delivery task~\cite{thomason2020jointly,veloso2018increasingly}. 
A delivery is deemed successful only if both the dialog and navigation subtasks are successfully conducted. 
We have conducted experiments using a simulated mobile robot, as well as demonstrated the system using a real mobile robot. 
Results show that the robot is able to learn world dynamics from navigation tasks through model-based RL, and apply the learned knowledge to both navigation tasks (with different goals) and delivery tasks (that require subtasks of navigation and dialog) through logical-probabilistic reasoning. 
In particular, we observed that the robot is able to adjust its dialog strategy through learning from navigation behaviors. 


\section{Related Work}
\label{sec:related}

Research areas related to this work include integrated logical KRR and RL, relational RL, and integrated KRR and probabilistic planning. 

Logical KRR has previously been integrated with RL. 
Action knowledge~\cite{mcdermott1998pddl,jiang2019task} has been used to reason about action sequences and help an RL agent explore only the states that can potentially contribute to achieving the ultimate goal~\cite{leonetti2016synthesis}. 
As a result, their agents learn faster by avoiding choosing ``unreasonable'' actions. 
A similar idea has been applied to domains with non-stationary dynamics~\cite{ferreira2017answer}. 
More recently, task planning was used to interact with the high level of a hierarchical RL framework~\cite{yang2018peorl}. 
The goal shared by these works is to enable RL agents to use knowledge to improve the performance in learning (e.g., to learn faster and/or avoid risky exploration). 
However, the KRR capabilities of these methods are limited to \emph{logical} action knowledge. 
By contrast, we use a logical-probabilistic KRR paradigm that can directly reason with probabilities learned from RL. 

Relational RL (RRL) combines RL with relational reasoning~\cite{dvzeroski2001relational}. 
Action models have been incorporated into RRL, resulting in a relational temporal difference learning method~\cite{asgharbeygi2006relational}. 
Recently, RRL has been deployed for learning affordance relations that forbid the execution of specific actions~\cite{sridharan2017can}. 
These RRL methods, including deep RRL~\cite{zambaldi2018relational}, exploit structural representations over states and actions in (only) current tasks. 
In this research, KRR-RL supports the KRR of world factors beyond those in state and action representations, e.g., \emph{time} in navigation tasks, as detailed in Section~\ref{sec:ins}. 

The research area of integrated KRR and probabilistic planning is related to this research. 
Logical-probabilistic reasoning has been used to compute informative priors and world dynamics~\cite{zhang2017dynamically,amiri2020learning} for probabilistic planning. 
An action language was used to compute a deterministic sequence of actions for robots, where individual actions are then implemented using probabilistic controllers~\cite{sridharan2019reba}. 
Recently, human-provided information has been incorporated into belief state representations to guide robot action selection~\cite{chitnis2018integrating}. 
In comparison to our approach, learning (from reinforcement or not) was not discussed in the above-mentioned algorithms. 

Finally, there are a number of robot reasoning and learning architectures~\cite{tenorth2013knowrob,oh2015toward,hanheide2017robot,khandelwal2017bwibots}, which are relatively complex, and support a variety of functionalities. 
In comparison, we aim at a concise representation for robot KRR and RL capabilities. 
To the best of our knowledge, this is the first work on a tightly coupled integration of logical-probabilistic KRR with model-based RL.

\section{Background}
\label{sec:background}

We briefly describe the two most important building blocks of this research, namely model-based RL and hybrid KRR.


\subsection{Model-based Reinforcement Learning}
\label{sec:pomdps}

Following the Markov assumption, a Markov decision process (MDP) can be described as a four-tuple $\langle \mathcal{S}, \mathcal{A}, T, R \rangle$~\cite{puterman1994markov}.
$\mathcal{S}$ defines the state set, where we assume a factored space in this work.
$\mathcal{A}$ is the action set.
$T:\mathcal{S}\times \mathcal{A}\times \mathcal{S}\rightarrow [0,1]$ specifies the state transition probabilities.
$R:\mathcal{S}\times \mathcal{A}\rightarrow \mathbb{R}$ specifies the rewards.
Solving an MDP produces an \emph{action policy} $\pi:s\mapsto a$ that maps a state to an action to maximize long-term rewards. 

RL methods fall into classes including model-based and model-free.
Model-based RL methods learn a model of the domain by approximating $R(s,a)$ and $P(s'|s,a)$ for state-action pairs, where $P$ represents the probabilistic transition system. An agent can then use planning methods to calculate an action policy~\cite{sutton1990integrated,kocsis2006bandit}.
Model-based methods are particularly attractive in this work, because they output partial world models that can better accommodate the diversity of tasks we are concerned with, c.f., model-free RL that is typically goal-directed. 

One of the best known examples of model-based RL is R-Max~\cite{brafman2002r}, which is guaranteed to learn a near-optimal policy with a polynomial number of suboptimal (exploratory) actions.
The algorithm classifies each state-action pair as known or unknown, according to the number of times it was visited.
When planning on the model, known state-actions are modeled with the learned reward, while unknown state-actions are given the maximum one-step reward, $R_{max}$.
This ``maximum-reward'' strategy automatically enables the agent to balance the exploration of unknown states and exploitation.
We use R-Max in this work, though KRR-RL practitioners can use supervised machine learning methods, e.g., imitation learning~\cite{osa2018algorithmic}, to build the model learning component.


\subsection{Logical Probabilistic KRR}
\label{sec:asp}

KRR paradigms are concerned with concisely representing and robustly reasoning with declarative knowledge.
Answer set programming (ASP) is a non-monotonic logical KRR paradigm~\cite{baral2010knowledge,gelfond2014knowledge} building on the stable model semantics~\cite{Gelfond:iclp88}.
An ASP program consists of a set of logical rules, in the form of ``\c{head :- body}'', that read ``\c{head} is true if \c{body} is true''.
Each ASP rule is of the form:
\begin{quote}
\begin{scriptsize}
\begin{verbatim}
a or ... or b :- c, ..., d, not e, ..., not f.
\end{verbatim}
\end{scriptsize}
\end{quote}
where \c{a...f} are literals that correspond to true or false statements. Symbol \c{not} is a logical connective called {\em default negation}; \c{not l} is read as ``it is not believed
that \c{l} is true'', which does not imply that \c{l} is false.
ASP has a variety of applications~\cite{erdem2016applications}.

Traditionally, ASP does not explicitly quantify degrees of uncertainty: a literal is either true, false or unknown.
P-log extends ASP to allow \emph{probability atoms} (or {\em pr-atoms})~\cite{baral2009probabilistic,balai2017refining}. The following pr-atom states that, if \c{B} holds, the probability of \c{a(t)=y} is \c{v}:
\begin{quote}
\begin{scriptsize}
\begin{verbatim}
pr(a(t)=y|B)=v.
\end{verbatim}
\end{scriptsize}
\end{quote}
where \c{B} is a collection of literals or their default negations;
\c{a} is a random variable;
\c{t} is a vector of terms (a term is a constant or a variable);
\c{y} is a term; and ${\tt v\in [0,1]}$.
Reasoning with an ASP program generates a set of {\em possible worlds}:
$\{W_0,W_1,\cdots\}$.
The pr-atoms in P-log enable calculating a probability for each possible world.
Therefore, P-log is a KRR paradigm that supports both logical and probabilistic inferences.
We use P-log in this work for KRR purposes.


\section{KRR-RL Framework}
\label{sec:alg}

\begin{figure}[tb]
  \vspace{-.1em}
  \begin{center}
    \includegraphics[width=0.5\textwidth]{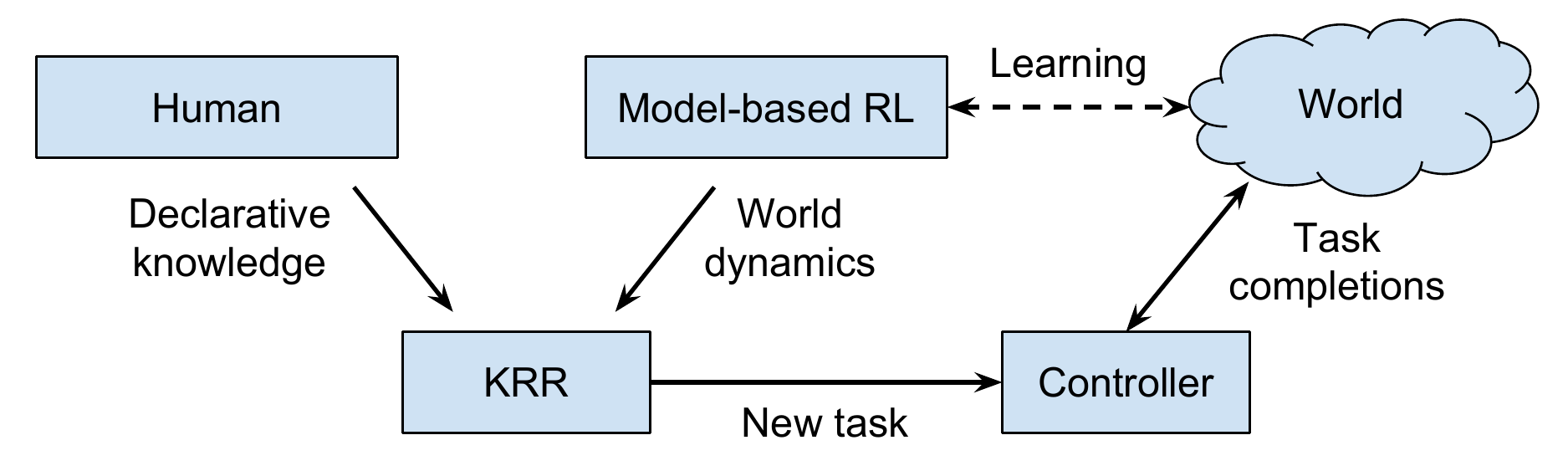}
  \end{center}
  \vspace{-1em}
  \caption{An overview of KRR-RL for robot learning and reasoning to complete complex tasks. }
  \label{fig:overview}
  \vspace{-1em}
\end{figure}

KRR-RL integrates logical-probabilistic KRR and model-based RL, and is illustrated in Figure~\ref{fig:overview}. 
The KRR component includes both declarative qualitative knowledge from humans and the probabilistic knowledge from model-based RL. 
When the robot is free, the robot arbitrarily selects goals (different navigation goals in our case) to work on, and learns the world dynamics, e.g., success rates and costs of navigation actions. 
When a task becomes available, the KRR component dynamically constructs a partial world model (excluding unrelated factors), on which a task-oriented controller is computed using planning algorithms. 
\textbf{Human knowledge concerns environment variables and their dependencies}, i.e., what variables are related to each action.
For instance, the human provides knowledge that navigation actions' success rates depend on current time and area (say elevator areas are busy in the mornings), \textbf{while the robot must learn specific probabilities} by interacting with the environment. 

\vspace{.5em}
\noindent
\textbf{\emph{Why is KRR-RL needed? }}	
Consider an indoor robot navigation domain, where a robot wants to maximize the success rate of moving to goal positions through navigation actions.
\emph{Shall we include factors, such as time, weather, positions of human walkers, etc, into the state space?}
On the one hand, to ensure model completeness, the answer should be ``yes''.
Human walkers and sunlight (that blinds robot's LiDAR sensors) reduce the success rates of the robot's navigation actions, 
and both can cause the robot irrecoverably lost.
On the other hand, to ensure computational feasibility, the answer is ``no''.
Modeling whether one specific grid cell being occupied by humans or not introduces one extra dimension in the state space, and doubles the state space size.
If we consider (only) ten such grid cells, the state space becomes $2^{10}\approx 1000$  times bigger.
As a result, RL practitioners frequently have to make a trade-off between model completeness and computational feasibility. 
In this work, we aim at a framework that retains both model scalability and computational feasibility, i.e., the agent is able to learn within relatively little memory while computing action policies accounting for a large number of domain variables.

\subsection{A General Procedure}

In factored spaces, state variables $\mathcal{V}=\{V_0,V_1,...,V_{n-1}\}$ can be split into two categories, namely endogenous variables $\mathcal{V}^{en}$ and exogenous variables $\mathcal{V}^{ex}$~\cite{chermack2004improving}, where $\mathcal{V}^{en}=\{V_0^{en},V_1^{en},...,V_{p-1}^{en}\}$ and $\mathcal{V}^{ex}=\{V_0^{ex},V_1^{ex},...,V_{q-1}^{ex}\}$. 
In our integrated KRR-RL context, $\mathcal{V}^{en}$ is goal-oriented and includes the variables whose values the robot wants to actively change so as to achieve the goal; and $\mathcal{V}^{ex}$ corresponds to the variables whose values affect the robot's action outcomes, but the robot cannot (or does not want to) change their values. 
Therefore, $\mathcal{V}^{en}$ and $\mathcal{V}^{ex}$ both depend on task $\tau$. 
Continuing the navigation example, robot position is an endogenous variable, and current time is an exogenous variable. 
For each task, $\mathcal{V}=\mathcal{V}^{en} \cup \mathcal{V}^{ex}$ and $n=p+q$, and RL agents learn in spaces specified by $\mathcal{V}^{en}$. 

The KRR component models $V$, their dependencies from human knowledge, and conditional probabilities on how actions change their values, as learned through model-based RL. 
When a task arrives, the KRR component uses probabilistic rules to generate a task-oriented Markov decision process (MDP)~\cite{puterman1994markov}, which only contains a subset of $\mathcal{V}$ that are relevant to the current task, i.e., $\mathcal{V}^{en}$, and their transition probabilities. 
Given this task-oriented MDP, a corresponding action policy is computed using value iteration or policy iteration.

\begin{algorithm}[tb]\footnotesize
  \caption{Learning in KRR-RL Framework}
  \label{alg:framework}
  \begin{algorithmic}[1]
    \REQUIRE{Logical rules $\Pi^L$; 
        probabilistic rules $\Pi^P$;
        random variables $\mathcal{V}=\{V_0,V_1,...,V_{n-1}\}$; 
        task selector $\Delta$; and 
        guidance functions (from human knowledge) of $f^V(\mathcal{V}, \tau)$ and $f^A(\tau)$}
    \WHILE{Robot has no task}
        \STATE{$\tau \leftarrow \Delta()$: a task is heuristically selected}
        \STATE{$\mathcal{V}^{en} \leftarrow f^V(\mathcal{V}, \tau)$, and $\mathcal{V}^{ex} \leftarrow \mathcal{V} \setminus \mathcal{V}^{en}$}
    \STATE{$A \leftarrow f^A(\tau)$}
    \STATE{$\mathcal{M} \leftarrow \textit{Procedure-}\ref{alg:generatemodel}(\Pi^L, \Pi^P, \mathcal{V}^{en}, \mathcal{V}^{ex}, A)$}
    \STATE{Initialize agent: $agent\leftarrow$ \emph{R-Max}$(\mathcal{M})$}
    \STATE{RL \emph{agent} repeatedly works on task $\tau$, and keeps maintaining task model $\mathcal{M}'$, until policy convergence}
    \ENDWHILE
    \STATE{Use $\mathcal{M}'$ to update $\Pi^P$}
  \end{algorithmic}
\end{algorithm}

Procedures~\ref{alg:framework} and \ref{alg:generatemodel} focus on how our KRR-RL agent learns by interacting with an environment when there is no task assigned.\footnote{As soon as the robot's learning process is interrupted by the arrival of a real service task (identified via dialog), it will call Procedure~\ref{alg:generatemodel} to generate a controller to complete the task. This process is not included in the procedures. } 
Next, we present the details of these two interleaved processes. 

Procedure~\ref{alg:framework} includes the steps of the learning process. 
When the robot is free, it interacts with the environment by heuristically selecting a task\footnote{Here curriculum learning in RL~\cite{narvekar2017autonomous} can play a role to task selection and we leave this aspect of the problem for future work.}, and repeatedly using a model-based RL approach, R-Max~\cite{brafman2002r} in our case, to complete the task. 
The two guidance functions come from human knowledge. 
For instance, given a navigation task, it comes from human knowledge that the robot should model its own position (specified by $f^V$) and actions that help the robot move between positions (specified by $f^A$). 
After the policy converges or this learning process is interrupted (e.g., by task arrivals), the robot uses the learned probabilities to update the corresponding world dynamics in KRR. 
For instance, the robot may have learned the probability and cost of navigating through a particular area in early morning. 
In case this learning process is interrupted, the so-far-``known'' probabilities are used for knowledge base update. 



\begin{figure*}[tb]
  \begin{center}
    \includegraphics[width=0.95\textwidth]{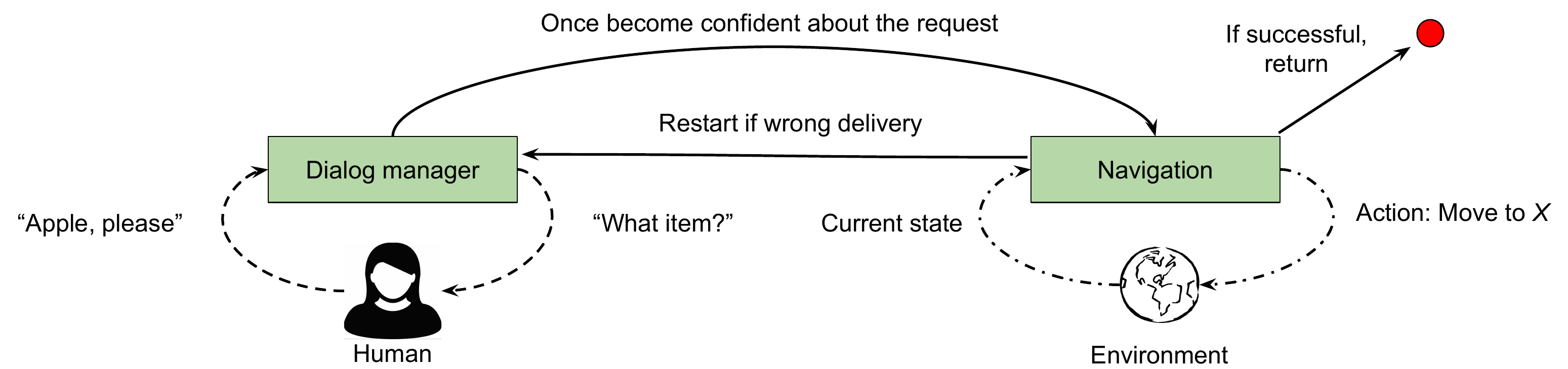}
  \end{center}
  \vspace{-1em}
  \caption{Transition system specified for delivery tasks, where question-asking actions are used for estimating the service request in dialog. Once the robot becomes confident about the service request, it starts to work on the navigation subtask. After the robot arrives, the robot might have to come back to the dialog subtask and redeliver, depending on whether the service request was correctly identified. }
  \label{fig:transition}
  \vspace{-1em}
\end{figure*}

Procedure~\ref{alg:generatemodel} includes the steps for building the probabilistic transition system of MDPs. 
The key point is that we consider only endogenous variables in the task-specific state space. 
However, when reasoning to compute the transition probabilities (Line~\ref{line:m}), the KRR component uses both $\Pi^P$ and $\mathcal{V}^{ex}$. 
The computed probabilistic transition systems are used for building task-oriented controllers, i.e., $\pi$, for task completions. 
In this way, the dynamically constructed controllers do not directly include exogenous variables, but their parameters already account for the values of all variables. 

Next, we demonstrate how our KRR-RL framework is instantiated on a real robot.


\subsection{An Instantiation on a Mobile Robot}
\label{sec:ins}
We consider a mobile service robot domain where a robot can do navigation, dialog, and delivery tasks. 
A \textbf{navigation task} requires the robot to use a sequence of (unreliable) navigation actions to move from one point to another. 
In a \textbf{dialog task}, the robot uses spoken dialog actions to specify service requests from people under imperfect language understanding. 
There is the trend of integrating language and navigation in the NLP and CV communities~\cite{chen2019touchdown,shridhar:cvpr20}. 
In this paper, they are integrated into \textbf{delivery tasks} that require the robot to use dialog to figure out the delivery request and conduct navigation tasks to physically fulfill the request. 
Specifically, a delivery task requires the robot to deliver item \c{I} to room \c{R} for person \c{P}, resulting in services in the form of \c{<I,R,P>}.
The challenges come from unreliable human language understanding (e.g., speech recognition) and unforeseen obstacles that probabilistically block the robot in navigation.

\begin{algorithm}[t]\footnotesize
  \caption{Model Construction for Task Completion}
  \label{alg:generatemodel}
  \begin{algorithmic}[1]
    \REQUIRE{$\Pi^L$; $\Pi^P$; $\mathcal{V}^{en}$; $\mathcal{V}^{ex}$; Action set $A$}
    \FOR{$V_i \in \mathcal{V}^{en}$, $i~$ in $[0,\cdots,|\mathcal{V}^{en}|\!-\!1]$}
        \FOR{each possible value $v$ in $range(V_i)$}
            \FOR{each $a\in A$}
                \FOR{each possible value $v'$ in $range(V_i)$}
                    \STATE{$\mathcal{M}(v'|a,v) \leftarrow$ Reason with $\Pi^L$ and $\Pi^P$} w.r.t $\mathcal{V}^{ex}$ \label{line:m}
                \ENDFOR
            \ENDFOR
        \ENDFOR
    \ENDFOR
    \RETURN $\mathcal{M}$
  \end{algorithmic}
\end{algorithm}



\paragraph{Human-Robot Dialog}

The robot needs spoken dialog to identify the request under unreliable language understanding, and navigation controllers for physically making the delivery.

The service request is not directly observable to the robot, and has to be estimated by asking questions, such as ``What item do you want?'' and ``Is this delivery for Alice?''
Once the robot is confident about the request, it takes a delivery action (i.e., \c{serve(I,R,P)}).
We follow a standard way to use partially observable MDPs (POMDPs)~\cite{kaelbling1998planning} to build our dialog manager, as reviewed in~\cite{Young2013POMDP}. 
The state set $\mathcal{S}$ is specified using \c{curr\_s}. 
The action set $\mathcal{A}$ is specified using \c{serve} and question-asking actions. 
Question-asking actions do not change the current state, and delivery actions lead to one of the terminal states (success or failure).
\footnote{More details in the supplementary document. }

After the robot becomes confident about the request via dialog, it will take a delivery action \c{serve\{I,R,P\}}. This delivery action is then implemented with a sequence of \c{act\_move} actions. When the request identification is incorrect, the robot needs to come back to the shop, figure out the correct request, and redeliver, where we assume the robot will correctly identify the request in the second dialog.
We use an MDP to model this robot navigation task, where the states and actions are specified using sorts \c{cell} and \c{move}. We use pr-atoms to represent the success rates of the unreliable movements, which are learned through model-based RL.
The dialog system builds on our previous work~\cite{lu2017leveraging}. 
Figure~\ref{fig:transition} shows the probabilistic transitions in delivery tasks.

\paragraph{Learning from Navigation}

We use R-Max~\cite{brafman2002r}, a model-based RL algorithm, to help our robot learn the success rate of navigation actions in different positions.
The agent first initializes an MDP, from which it uses R-Max to learn the partial world model (of navigation tasks). Specifically, it initializes the transition function with $T^N(s,a,s^v)=1.0$, where $s\in \mathcal{S}$ and $a\in \mathcal{A}$, meaning that starting from any state, after any action, the next state is always $s^v$. The reward function is initialized with $\mathcal{R}(s,a)=R_{max}$, where $R_{max}$ is an upper bound of reward. The initialization of $T^N$ and $\mathcal{R}$ enables the learner to automatically balance exploration and exploitation.
There is a fixed small cost for each navigation action. The robot receives a big bonus if it successfully achieves the goal ($R_{max}$), whereas it receives a big penalty otherwise ($-R_{max}$).
A transition probability in navigation, $T^N(s,a,s')$, is not computed until there are a minimum number ($M$) of transition samples visiting $s'$.
We recompute the action policy after $E$ action steps.

\paragraph{Dialog-Navigation Connection}
The update of knowledge base is achieved through updating the success rate of delivery actions \c{serve(I,R,P)} (in dialog task) using the success rate of navigation actions \c{act\_move=M} in different positions.
\begin{align*}
\label{eqn:transfer}
    &T^D(s^r, a^d, s^t)= \nonumber\\
    &\!\!\!\begin{cases}
        P^N(s^{sp}\!, s^{gl}),~{\bf if}~s^r\odot a^d\\
        P^N(s^{sp}\!, s^{mi})\!\times\! P^N(s^{mi}\!, s^{sp}) \!\times\! P^N(s^{sp}\!, s^{gl}),~{\bf if}~s^r\!\otimes a^d
    \end{cases}
\end{align*}
where $T^D(s^r, a^d, s^t)$ is the probability of fulfilling request $s^r$ using delivery action $a^d$;
$s^t$ is the ``success'' terminal state;
$s^{sp}$, $s^{mi}$ and $s^{gl}$ are states of the robot being in the shop, a misidentified goal position, and real goal position respectively;
and $P^N(s,s')$ is the probability of the robot successfully navigating from $s$ to $s'$ positions. 
When $s^r$ and $a^d$ are aligned in all three dimensions (i.e., $s^r\odot a^d$), the robot needs to navigate once from the shop ($s^{sp}$) to the requested navigation goal ($s^{gl}$).
$P^N(s^{sp},s^{gl})$ is the probability of the corresponding navigation task. When the request and delivery action are not aligned in at least one dimension (i.e., $s^{r}\otimes a^{d}$), the robot has to navigate back to the shop to figure out the correct request, and then redeliver, resulting in three navigation tasks.

\emph{Intuitively, the penalty of failures in a dialog subtask depends on the difficulty of the wrongly identified navigation subtask. }
For instance, a robot supposed to deliver to a near (distant) location being wrongly directed to a distant (near) location, due to a failure in the dialog subtask, will produce a higher (lower) penalty to the dialog agent.

\begin{figure}[tb]
 \vspace{.5em}
  \begin{center}
    \includegraphics[height=0.14\textheight]{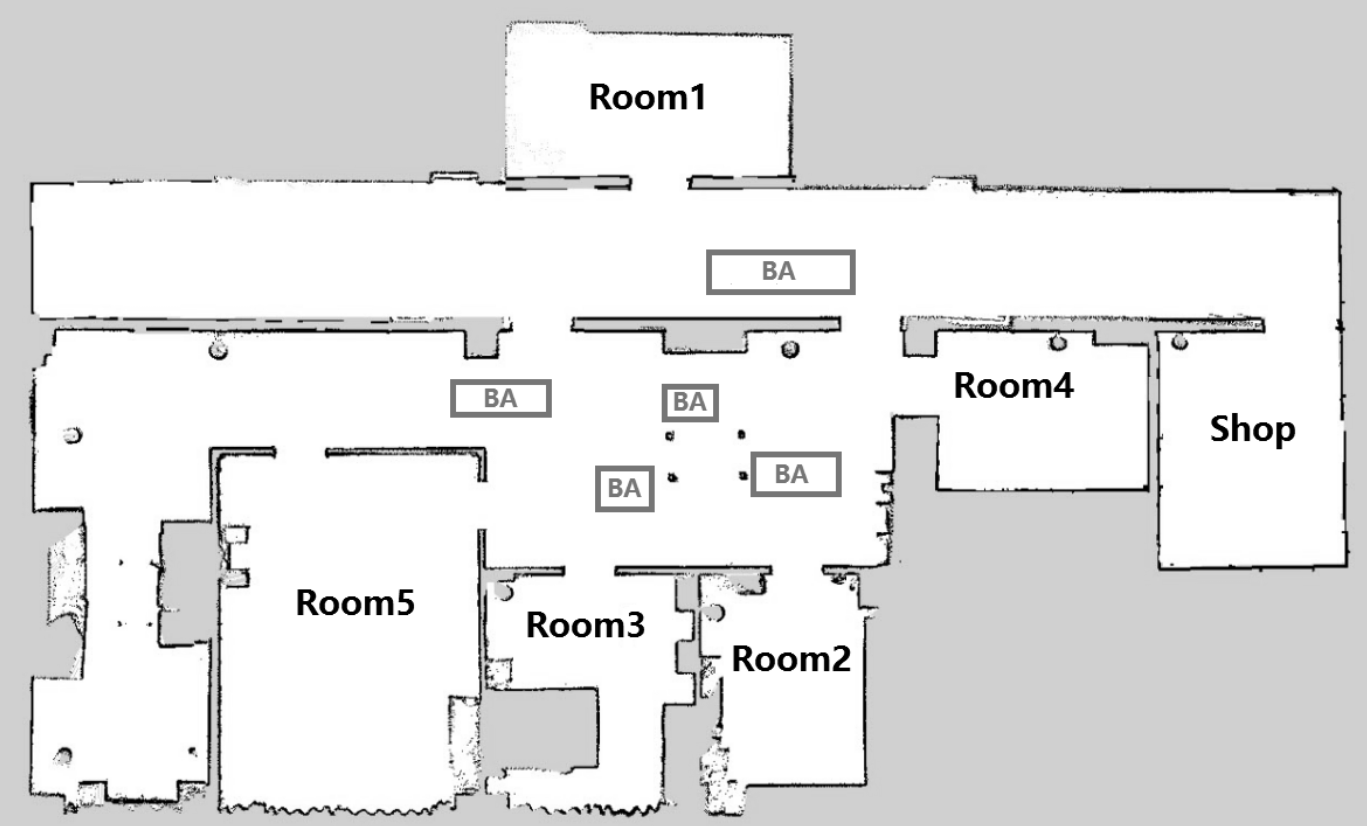}
    \includegraphics[height=0.14\textheight]{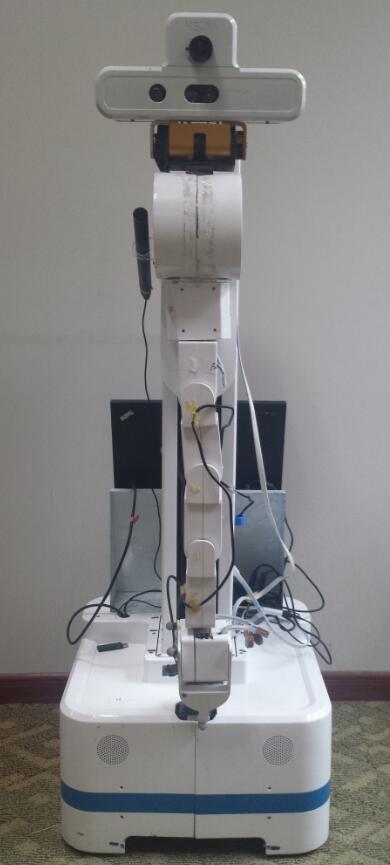}
  \end{center}
  \vspace{-.5em}
  \caption{Occupancy-grid map used in our experiments (\textbf{Left}), including five rooms, one shop, and four blocking areas (indicated by `BA'), where all deliveries are from the shop and to one of the rooms; and (\textbf{Right}) mobile robot platform used in this research.}
  \label{fig:map}
  \vspace{-1em}
\end{figure}

\begin{figure}[tb]
 \vspace{.5em}
  \begin{center}
    \includegraphics[width=0.48\columnwidth]{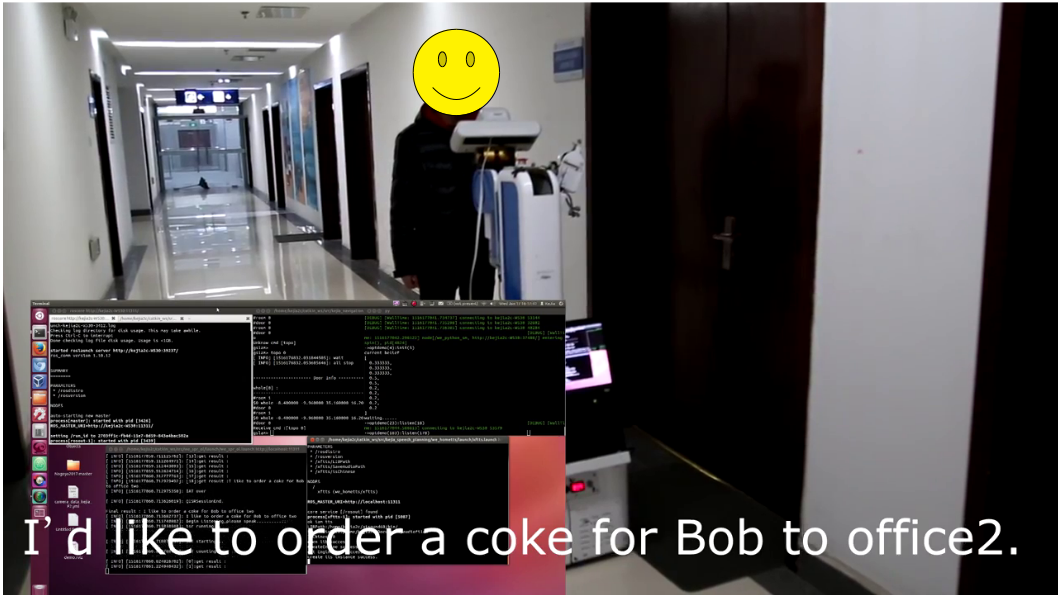}
    \includegraphics[width=0.48\columnwidth]{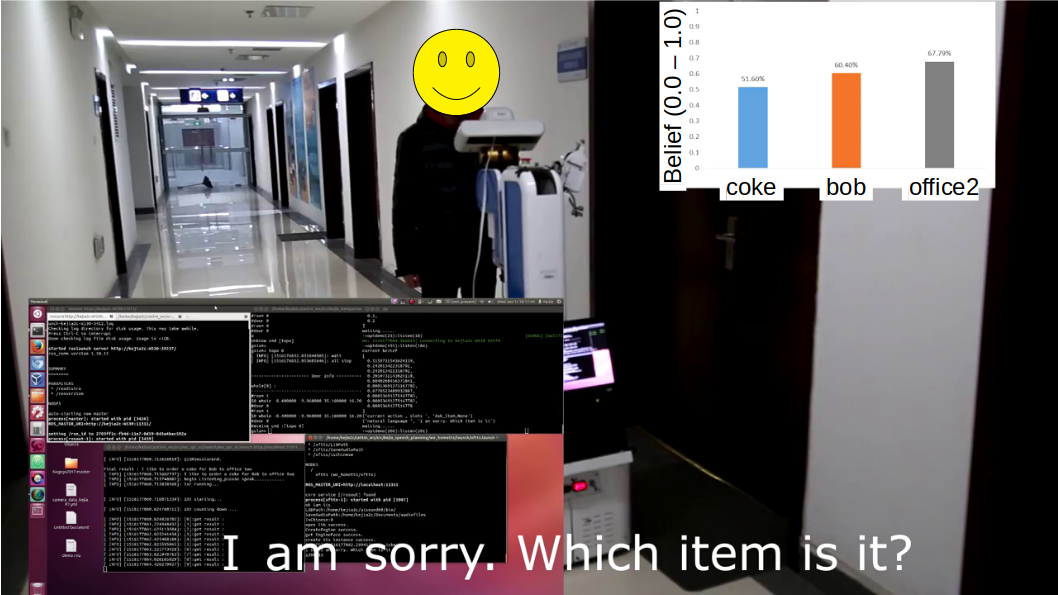}\\
    {\small (a)~~~~~~~~~~~~~~~~~~~~~~~~~~~~~~~~~~~~~~~~~~~~~~~~~(b)}\\ \vspace{.5em}
    \includegraphics[width=0.48\columnwidth]{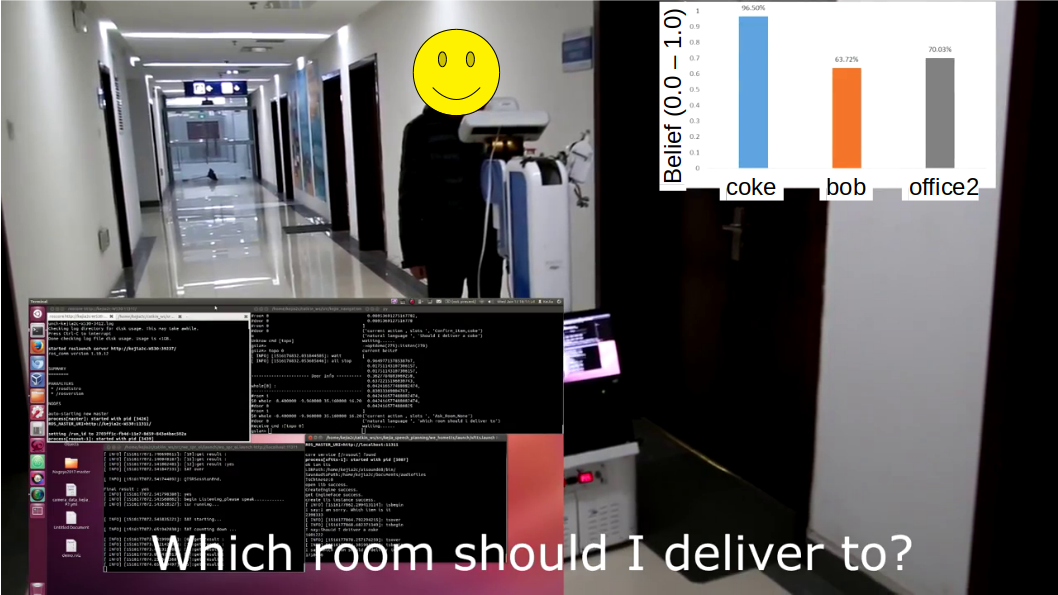}
    \includegraphics[width=0.48\columnwidth]{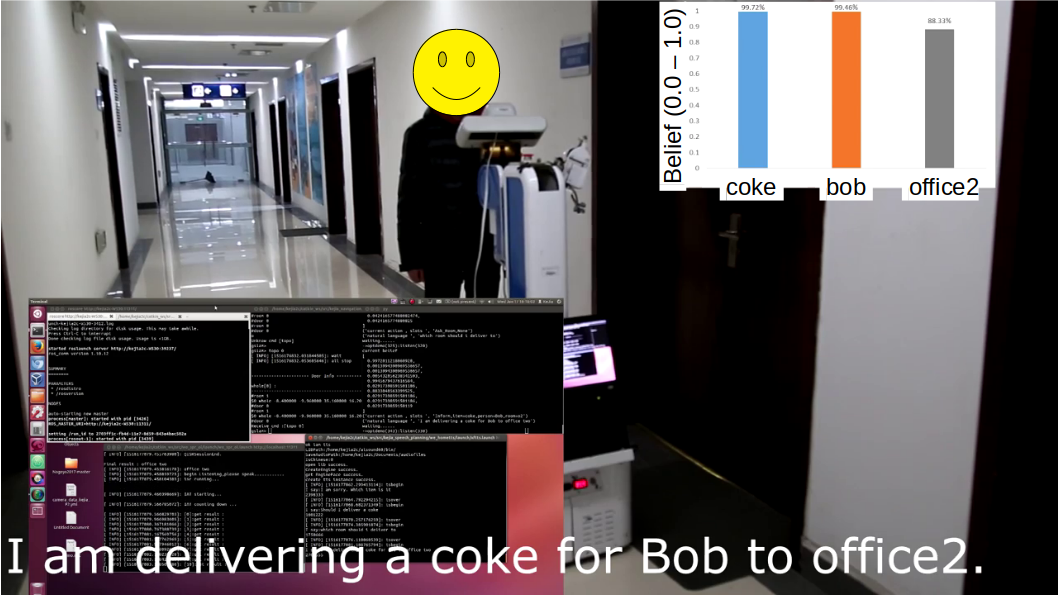}\\
    {\small (c)~~~~~~~~~~~~~~~~~~~~~~~~~~~~~~~~~~~~~~~~~~~~~~~~~(d)}\\ \vspace{.5em}
    \includegraphics[width=0.48\columnwidth]{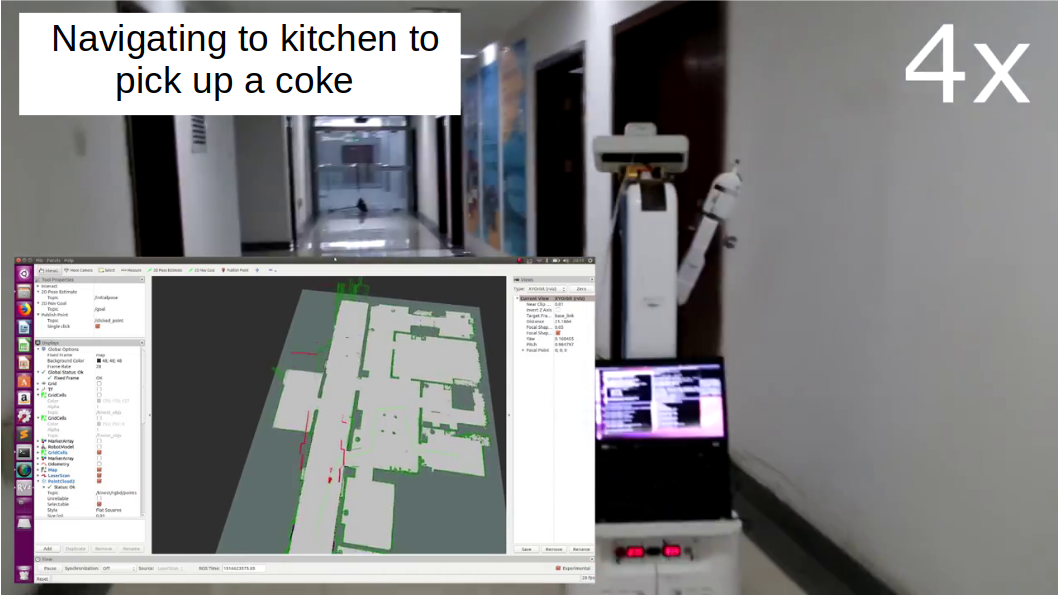}
    \includegraphics[width=0.48\columnwidth]{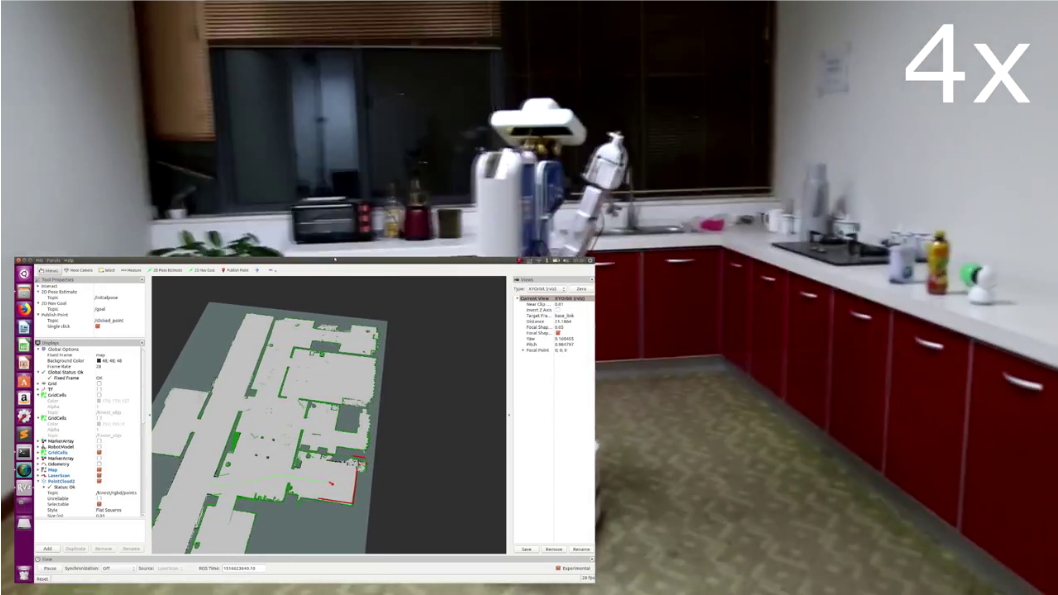}\\
    {\small (e)~~~~~~~~~~~~~~~~~~~~~~~~~~~~~~~~~~~~~~~~~~~~~~~~~(f)}
  \end{center}
  \vspace{-1em}
  \caption{Screenshots of a demonstration trial on a real robot. (a) User gives the service request; 
  (b) The robot decided to confirm about the item, considering its unreliable language understanding capability; 
  (c) After hearing ``coke'', the robot became more confident about the item, and decided to ask again about the goal room'; 
  (d) After hearing ``office2'', the robot became confident about the whole request, and started to work on the task; 
  (e) Robot was on the way to the kitchen to pick up the object; and 
  (f) Robot arrived at the kitchen, and was going to pick up the object for delivery. }
  \label{fig:demo}
  \vspace{-.5em}
\end{figure}

\begin{figure*}[t]
  \hspace*{-.5em}
  \begin{center}
    \includegraphics[width=0.8\textwidth]{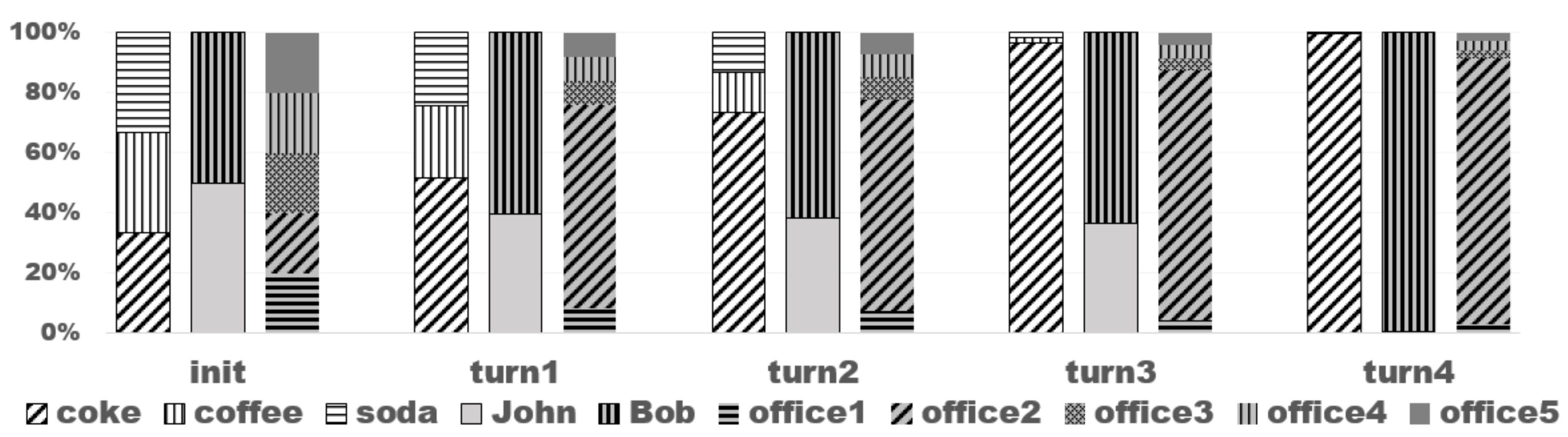}
  \end{center}
   \vspace{-.5em}
  \caption{Belief change in three dimensions (In order from the left: Items, Persons and Offices) over five turns in a human-robot dialog .
  The distributions are grouped by turns (Including the initial distribution). In each turn, there are three distribution bars which means three different dimensions (In order from the left: Item, Person and Office). In order from the bottom, the values in each dimension are 1) coke, coffee and soda in Item; 2) John and Bob in Person; and 3) office1, office2, office3, office4 and office5 in Office.}
  \label{fig:distribution}
  \vspace{-1.0em}
\end{figure*}


\section{Experiments}
\label{sec:experiment}


In this section, the goal is to evaluate our hypothesis that our KRR-RL framework enables a robot to learn from model-based RL, reason with both the learned knowledge and human knowledge, and dynamically construct task-oriented controllers.
Specifically, our robot learns from navigation tasks, and applied the learned knowledge (through KRR) to navigation, dialog, and delivery tasks. 

We also evaluated whether the learned knowledge can be represented and applied to tasks under different world settings. 
In addition to simulation experiments, we have used a real robot to demonstrate how our robot learns from navigation to perform better in dialog.
Figure~\ref{fig:map} shows the map of the working environment (generated using a real robot) used in both simulation and real-robot experiments. Human walkers in the blocking areas (``BA'') can probabilistically impede the robot, resulting in different success rates in navigation tasks.

We have implemented our KRR-RL framework on a mobile robot in an office environment.
As shown in Figure~\ref{fig:map}, 
the robot is equipped with two Lidar sensors for localization and obstacle avoidance in navigation, and a Kinect RGB-D camera for human-robot interaction.
We use the Speech Application Programming Interface (SAPI) package (\texttt{\small \url{http://www.iflytek.com/en}}) for speech recognition.
The robot software runs in the Robot Operating System (ROS)~\cite{quigley2009ros}.

\paragraph{An Illustrative Trial on a Robot:}

Figure~\ref{fig:demo} shows the screenshots of milestones of a demo video, which will be made available given its acceptance. 
After hearing ``a coke for Bob to office2'', the three sub-beliefs are updated (\c{turn1}).
Since the robot is aware of its unreliable speech recognition, it asked about the item, ``Which item is it?''
After hearing ``a coke'', the belief is updated (\c{turn2}), and the robot further confirmed on the item by asking ``Should I deliver a coke?''
It received a positive response (\c{turn3}), and decided to move on to ask about the delivery room: ``Should I deliver to office 2?''
\emph{After this question, the robot did not further confirm the delivery room, because it learned through model-based RL that navigating to \c{office2} is relatively easy and it decided that it is more worth risking an error and having to replan than it is to ask the person another question. }
The robot became confident in three dimensions of the service request (\c{<coke,Bob,office2>} in \c{turn4}) \emph{without} asking about \c{person}, because of the prior knowledge (encoded in P-log) about \c{Bob}'s office. 

Figure~\ref{fig:distribution} shows the belief changes (in the dimensions of \c{item}, \c{person}, and \c{room}) as the robot interacts with a human user.
The robot started with a uniform distribution in all three categories. It should be noted that, although the marginal distributions are uniform, the joint belief distribution is not, as the robot has prior knowledge such as \c{Bob}'s office is \c{office2} and people prefer deliveries to their own offices. 
Demo video is not included to respect the anonymous review process.





\begin{figure}[tb]
  \begin{center}
    \hspace*{-.1em}\includegraphics[width=0.26\textwidth]{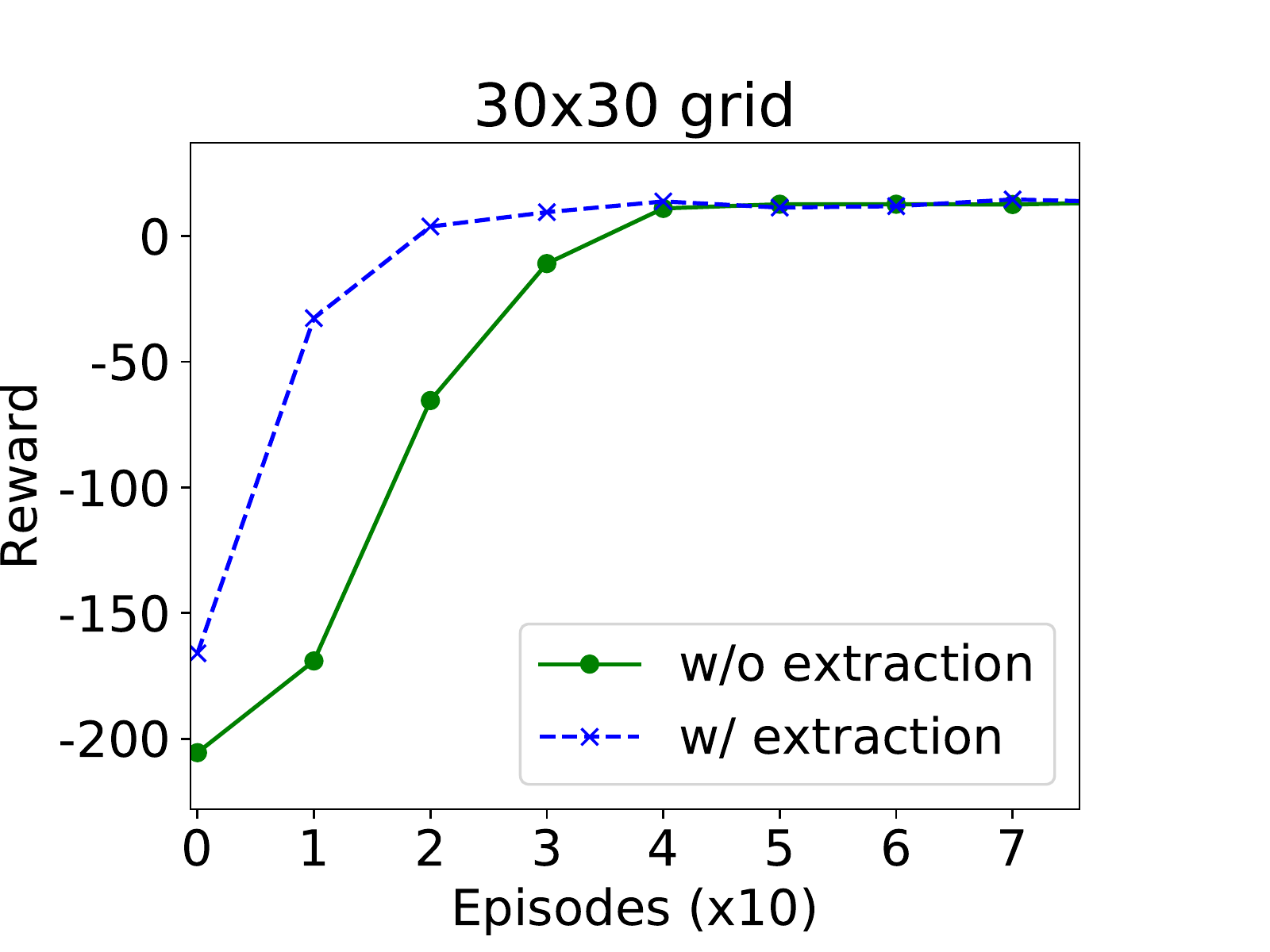} \hspace*{-1.7em}
    \includegraphics[width=0.26\textwidth]{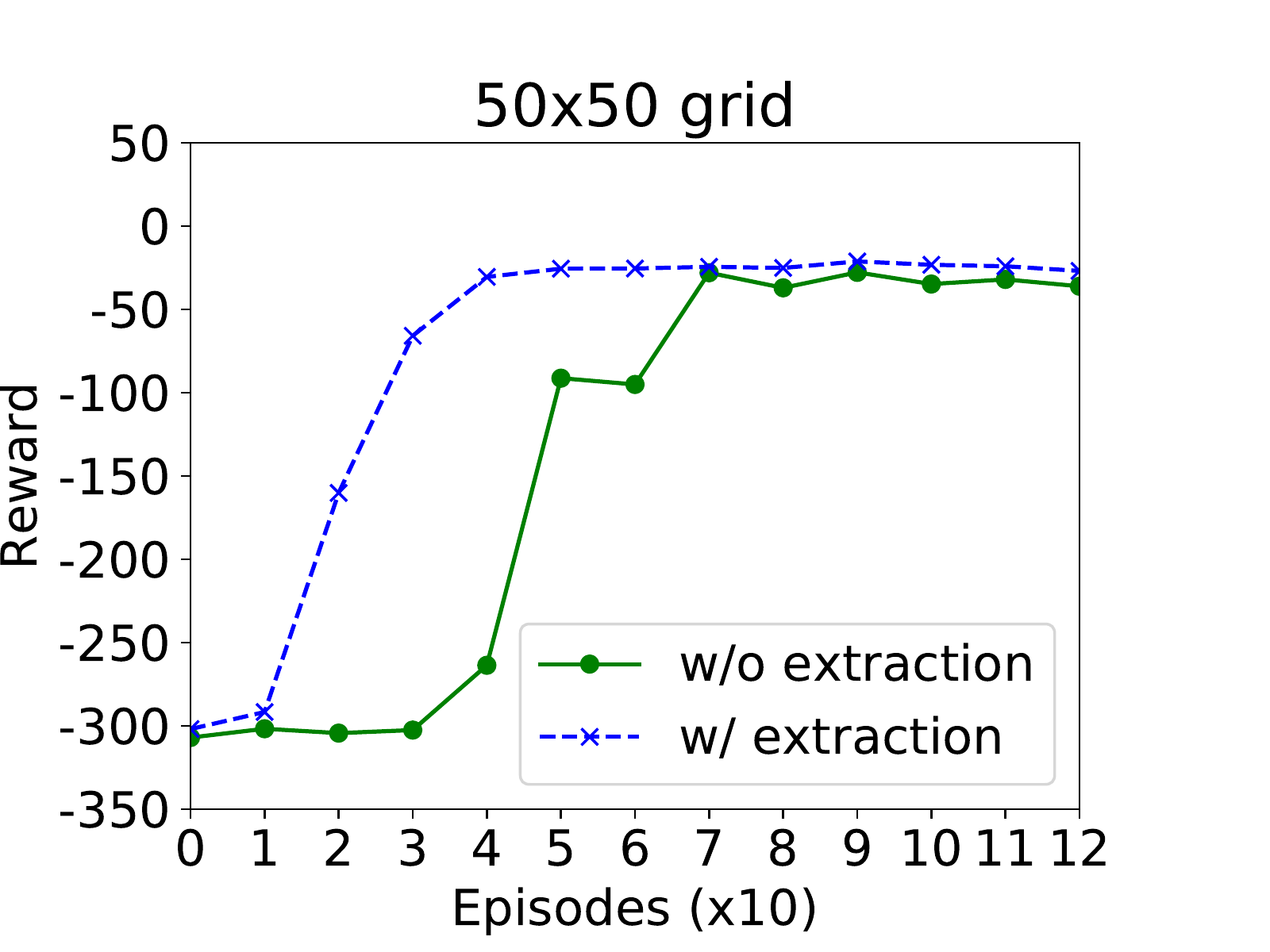}\hspace*{-1.4em}
  \end{center}
  \vspace{-1em}
  \caption{Navigation tasks in small (Left: $30\times 30$ grid) and large (Right: $50\times 50$ grid) domains.
  With extraction (KRR-RL in dashed line), the robot learns faster in the target navigation task. }
  \label{fig:navigation}
  \vspace{-1em}
\end{figure}

\paragraph{Learning to Navigate from Navigation Tasks}
In this experiment, the robot learns in the \c{shop-room1} navigation task, and extracts the learned partial world model to the \c{shop-room2} task. It should be noted that navigation from \c{shop} to \c{room2} requires traveling in areas that are unnecessary in the \c{shop-room1} task. 

Figure~\ref{fig:navigation} presents the results, where each data points corresponds to an average of 1000 trials. Each episode allows at most 200 (300) steps in small (large) domain. The curves are smoothed using a window of 10 episodes. The results suggest that with knowledge extraction (the dashed line) the robot learns faster than without extraction, and this performance improvement is more significant in a larger domain (the Right subfigure).

\paragraph{Learning to Dialog and Navigate from Navigation Tasks}

Robot delivering objects requires both tasks: dialog management for specifying service request (under unreliable speech recognition) and navigation for physically delivering objects (under unforeseen obstacles).
Our office domain includes five rooms, two persons, and three items, resulting in 30 possible service requests.
In the dialog manager, the reward function gives delivery actions a big bonus (80) if a request is fulfilled, and a big penalty (-80) otherwise.

General questions and confirming questions cost 2.0 and 1.5 respectively. In case a dialog does not end after 20 turns, the robot is forced to work on the most likely delivery. 
The cost/bonus/penalty values are heuristically set in this work, following guidelines based on studies from the literature on dialog agent behaviors~\cite{zhang2015corpp}. 

\renewcommand{\arraystretch}{1}
\begin{table}[h]
\vspace{.2em}
  \centering
  \caption{Overall performance in delivery tasks (requiring both dialog and navigation). }
  \label{tab:both} \vspace{-.5em}
  \fontsize{6}{8}\selectfont
  \begin{threeparttable}
    \begin{tabular}{ccccccc}
    \toprule
    \multirow{2}{*}{}&\multicolumn{3}{c}{Static policy}&\multicolumn{3}{c}{KRR-RL}\cr
    \cmidrule(lr){2-4} \cmidrule(lr){5-7}
    & Reward & Fulfilled & QA Cost & Reward & Fulfilled & QA Cost \cr
    \midrule
    $br=0.1$ &182.07 &0.851 &20.86     &206.21 &0.932 &18.73 \cr
    $br=0.5$ & 30.54 &0.853 &20.84     & 58.44 &0.927 &18.98 \cr
    $br=0.7$ &-40.33 &0.847 &20.94     &-14.50 &0.905 &20.56 \cr
    \bottomrule
    \end{tabular}
    \vspace{-.5em}
    \end{threeparttable}
\end{table}

Table~\ref{tab:both} reports the robot's overall performance in  delivery tasks, which requires \textbf{accurate dialog} for identifying delivery tasks and \textbf{safe navigation} for object delivery.
We conduct 10,000 simulation trials under each blocking rate.
Without learning from RL, the robot uses a world model (outdated) that was learned under $br=0.3$. With learning, the robot updates its world model in domains with different blocking rates.
We can see, when learning is enabled, our KRR-RL framework produces higher overall reward, higher request fulfillment rate, and lower question-asking cost. 
\emph{The improvement is statistically significant}, i.e., the $p\!-\!values$ are 0.028, 0.035, and 0.049 for overall reward, when \emph{br} is 0.1, 0.5, and 0.7 respectively (100 randomly selected trials with/without extraction).

\paragraph{Learning to Adjust Dialog Strategies from Navigation}
In the last experiment, we quantify the information collected in dialog in terms of entropy reduction. The hypothesis is that, using our KRR-RL framework, the dialog manager wants to collect more information before physically working on more challenging tasks.
In each trial, we randomly generate a belief distribution over all possible service requests, evaluate the entropy of this belief, and record the suggested action given this belief. We then statistically analyze the entropy values of beliefs, under which delivery actions are suggested.

\renewcommand{\arraystretch}{1}
\begin{table}[h]
  \centering
   \vspace{.5em}
  \caption{The amount of information (in terms of entropy) needed by a robot before taking delivery actions.}
  \vspace{-.5em}
  \label{tab:entropy}
  \fontsize{6}{8}\selectfont  
  \begin{threeparttable}
    \begin{tabular}{ccccccc}
    \toprule
    \multirow{3}{*}{}   &\multicolumn{2}{c}{Entropy (room1)}
                        &\multicolumn{2}{c}{Entropy (room2)}
                        &\multicolumn{2}{c}{Entropy (room5)}\cr
    \cmidrule(lr){2-3} \cmidrule(lr){4-5} \cmidrule(lr){6-7}
    & Mean (std) & Max & Mean (std) & Max & Mean (std) & Max \cr
    \midrule
    $br=0.1$ & .274 (.090) & .419 & .221 (.075)     & .334 & .177 (.063) & .269 \cr
    $br=0.5$ & .154 (.056) & .233 & .111 (.044)     & .176 & .100 (.041) & .156 \cr
    $br=0.7$ & .132 (.050) & .207 & .104 (.042)     & .166 & .100 (.041) & .156 \cr
    \bottomrule
    \end{tabular}
    \vspace{-1.5em}
    \end{threeparttable}
\end{table}

Table~\ref{tab:entropy} shows that, when \emph{br} grows from 0.1 to 0.7, the means of belief entropy decreases (i.e., belief is more converged).
This suggests that the robot collected more information in dialog in environments that are more challenging for navigation, which is consistent with Table~1 in the main paper.
Comparing the three columns of results, we find the robot collects the most information before it delivers to \c{room5}.
This is because such delivery tasks are the most difficult due to the location of \c{room5}. The results support our hypothesis that learning from navigation tasks enables the robot to adjust its information gathering strategy in dialog given tasks of different difficulties.

\paragraph{Adaptive Control in New Circumstances}
The knowledge learned through model-based RL is contributed to a knowledge base that can be used for many tasks. 
So our KRR-RL framework enables a robot to dynamically generate partial world models for tasks under settings that were never experienced. 
For example, an agent does not know the current time is morning or noon, there are two possible values for variable ``time''. 
Consider that our agent has learned world dynamics under the times of morning and noon. 
Our KRR-RL framework enables the robot to reason about the two transition systems under the two settings and generate a new transition system for this ``morning-or-noon'' setting. 
Without our framework, an agent would have to randomly select one between the ``morning'' and ``noon'' policies. 


\begin{figure}[tb]
  \begin{center}
    \includegraphics[width=0.5\columnwidth]{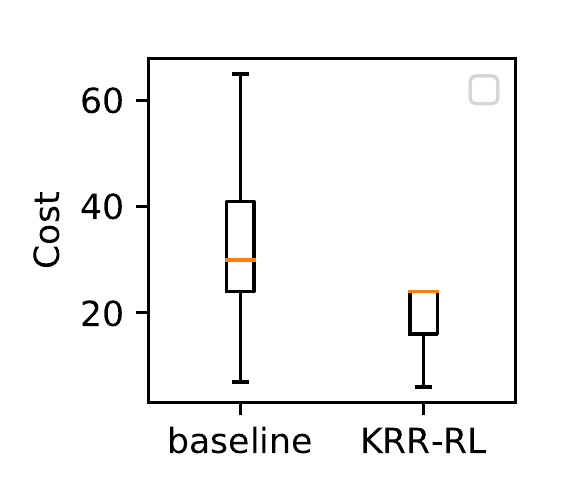}
  \end{center}
  \vspace{-1.5em}
  \caption{Adaptive behaviors under new circumstances.}
  \vspace{-1em}
  \label{fig:circumstances}
\end{figure}

To evaluate our policies dynamically constructed via KRR, we let an agent learn three controllers under three different environment settings -- the navigation actions have decreasing success rates under the settings. 
In this experiment, the robot does not know which setting it is in (out of two that are randomly selected). 
The baseline does not have the KRR capability of merging knowledge learned from different settings, and can only randomly select a policy from the two (each corresponding to a setting). 
Experimental results show that the baseline agent achieved an average of $26.8\%$ success rate in navigation tasks, whereas our KRR-RL agent achieved $83.8\%$ success rate on average. 
Figure~\ref{fig:circumstances} shows the costs in a box plot (including min-max, $25\%$, and $75\%$ values). 
Thus, KRR-RL enables a robot to effectively apply the learned knowledge to tasks under new settings. 

Let us take a closer look at the ``time'' variable $T$. 
If $\mathcal{T}$ is the domain of $T$, the RL-only baseline has to compute a total of $2^{|\mathcal{T}|}$ world models to account for all possible information about the value of $T$, where $2^{|\mathcal{T}|}$ is the number of subsets of $\mathcal{T}$. 
If there are $N$ such variables, the number of world models grows exponentially to $2^{|\mathcal{T}|\cdot N}$. 
In comparison, the KRR-RL agent needs to compute only $|\mathcal{T}|^N$ world models, which dramatically reduces the number of parameters that must be learned through RL while retaining policy quality.




\section{Conclusions and Future Work}
\label{sec:conclude}
We develop a KRR-RL framework that integrates computational paradigms of logical-probabilistic knowledge representation and reasoning (KRR), and model-based reinforcement learning (RL). 
Our KRR-RL agent learns world dynamics via model-based RL, and then incorporates the learned dynamics into the logical-probabilistic reasoning module, which is used for dynamic construction of efficient run-time task-specific planning models. 
Experiments were conducted using a mobile robot (simulated and physical) working on delivery tasks that involve both navigation and dialog. 
Results suggested that the learned knowledge from RL can be represented and used for reasoning by the KRR component, enabling the robot to dynamically generate task-oriented action policies. 


The integration of a KRR paradigm and model-based RL paves the way for  at least the following research directions. 
We plan to study how to sequence source tasks to help the robot perform the best in the target task (i.e., a curriculum learning problem within the RL context~\cite{narvekar2017autonomous}). 
Balancing the efficiencies between service task completion and RL is another topic for further study -- currently the robot optimizes for task completions (without considering the potential knowledge learned in this process) once a task becomes available. 
Fundamentally, all domain variables are endogenous, because one can hardly find variables whose values are completely independent from robot actions. 
However, for practical reasons (such as limited computational resources), people have to limit the number of endogenous. 
It remains an open question of how to decide what variables should be considered as being endogenous.

\section*{Acknowledgments}
{\small This work is supported in part by the National Natural Science Foundation of China under grant number U1613216. 
This work has taken place partly in the Autonomous Intelligent Robotics (AIR) Group at SUNY Binghamton. AIR research is supported in part by grants from the National Science Foundation (IIS-1925044), Ford Motor Company (URP Award), OPPO (Faculty Research Award), and SUNY Research Foundation.
This work has taken place partly in the Learning Agents Research Group (LARG) at the Artificial Intelligence Laboratory, The University of Texas at Austin. LARG research is supported in part by grants from the National Science Foundation (CPS-1739964, IIS-1724157, NRI-1925082), the Office of Naval Research (N00014-18-2243), Future of Life Institute (RFP2-000), DARPA, Lockheed Martin, General Motors, and Bosch. The views and conclusions contained in this document are those of the authors alone. Peter Stone serves as the Executive Director of Sony AI America and receives financial compensation for this work. The terms of this arrangement have been reviewed and approved by the University of Texas at Austin in accordance with its policy on objectivity in research.}

{
\bibliographystyle{acl_natbib}
\bibliography{references}
}

\clearpage

\appendix

\section{World Model for Dialog and Navigation}
We present complete world models constructed using P-log~\cite{baral2009probabilistic,balai2017refining}, a logical-probabilistic KRR paradigm, and describe the process of constructing task-oriented controllers.
It should be noted that the robot learns world dynamics and action costs using model-based RL; all other information is provided as prior knowledge.

\paragraph{Representing Rigid Knowledge}

Rigid knowledge includes information that does not depend upon the passage of time. We introduce a set of \emph{sorts} including \c{time}, \c{person}, \c{item}, and \c{room}, so as to specify a complete space of service requests. For instance, we can use \c{time=\{morning, noon, afternoon, ...\}} to specify possible values of \c{time}.



To model navigation domains, we introduce $N$ grid cells using \c{cell=\{0,...,N-1\}}, where the robot can travel. The geometric organization of these cells can be specified using predicates of \c{leftof($n_1$,$n_2$)} and \c{belowof($m_1$,$m_2$)}, where $n_i$ and $m_j$ are cells, $i,j\in \{0,\dots, N-1\}$.

We use a set of random variables to model the space of world states.
For instance, \c{curr\_cell:cell} states that the robot's position \c{curr\_cell}, as a random variable, must be in sort \c{cell};
\c{curr\_room:person->room} states that a person's current room must be in sort \c{room};
and \c{curr\_succ} identifies the request being fulfilled or not.
%
%
We can then use \c{random(curr\_cell)} to state that the robot's current position follows a uniform distribution over \c{cell}, unless specified elsewhere. 


\paragraph{Representing Action Knowledge}

We use a set of random variables in P-log to specify the set of delivery actions available to the robot. For instance, \c{serve(coffee,lab,alice)} indicates that the current service request is to deliver \c{coffee} to \c{lab} for \c{alice}.
\begin{quote}
\begin{scriptsize}
\begin{verbatim}
serve(I,R,P) :- act_item=I, act_room=R, 
                act_person=P.
\end{verbatim}
\end{scriptsize}
\end{quote}
where predicates with prefix \c{act\_} are random variables for modeling \c{serve} actions.

To model probabilistic transitions, we need one-step lookahead, modeling how actions lead state transitions. We introduce two identical state spaces using predicates \c{curr\_s} and \c{next\_s}. The following shows the specification of the current state using \c{curr\_s}.

\begin{quote}
\begin{scriptsize}
\begin{verbatim}
curr_s(I,R,P,C,S) :- curr_item(P)=I, 
            curr_room(P)=R, curr_person=P, 
            curr_cell=C, curr_succ=S.
\end{verbatim}
\end{scriptsize}
\end{quote}



Given the current and next state spaces, we can use pr-atoms to specify the transition probabilities led by delivery actions.
For instance, to model physical movements, we introduce random variable \c{act\_move:move} and sort \c{move=\{up, down, left, right\}}. Then we can use \c{act\_move=right} to indicate the robot attempting to move rightward by one cell, and use the following pr-atom
\begin{quote}
\begin{scriptsize}
\begin{verbatim}
pr(next_cell=C1 | curr_cell=C, leftof(C,C1),
                  act_move=right) = 8/10.
\end{verbatim}
\end{scriptsize}
\end{quote}
to state that, after taking action \c{right}, the probability of the robot successfully navigating to the cell on the right is $0.8$ (otherwise, it ends up with one of the nearby cells). 
\emph{Such probabilities are learned through model-based RL.}

It should be noted that, even if a request is correctly identified in dialog, the robot still cannot always succeed in delivery, because there are obstacles that can probabilistically trap the robot in navigation.
When the request is misidentified, delivery success rate drops, because the robot has to conduct multiple navigation tasks to figure out the correct request and redo the delivery.
We use $s\odot a$ and $s\otimes a$ to represent delivery action $a$ matches to request component of $s$ or not (i.e., service request is correctly identified in dialog or not).

\begin{figure}[tb]
  \begin{center}
    \hspace*{-.5em}
    \includegraphics[width=0.26\textwidth]{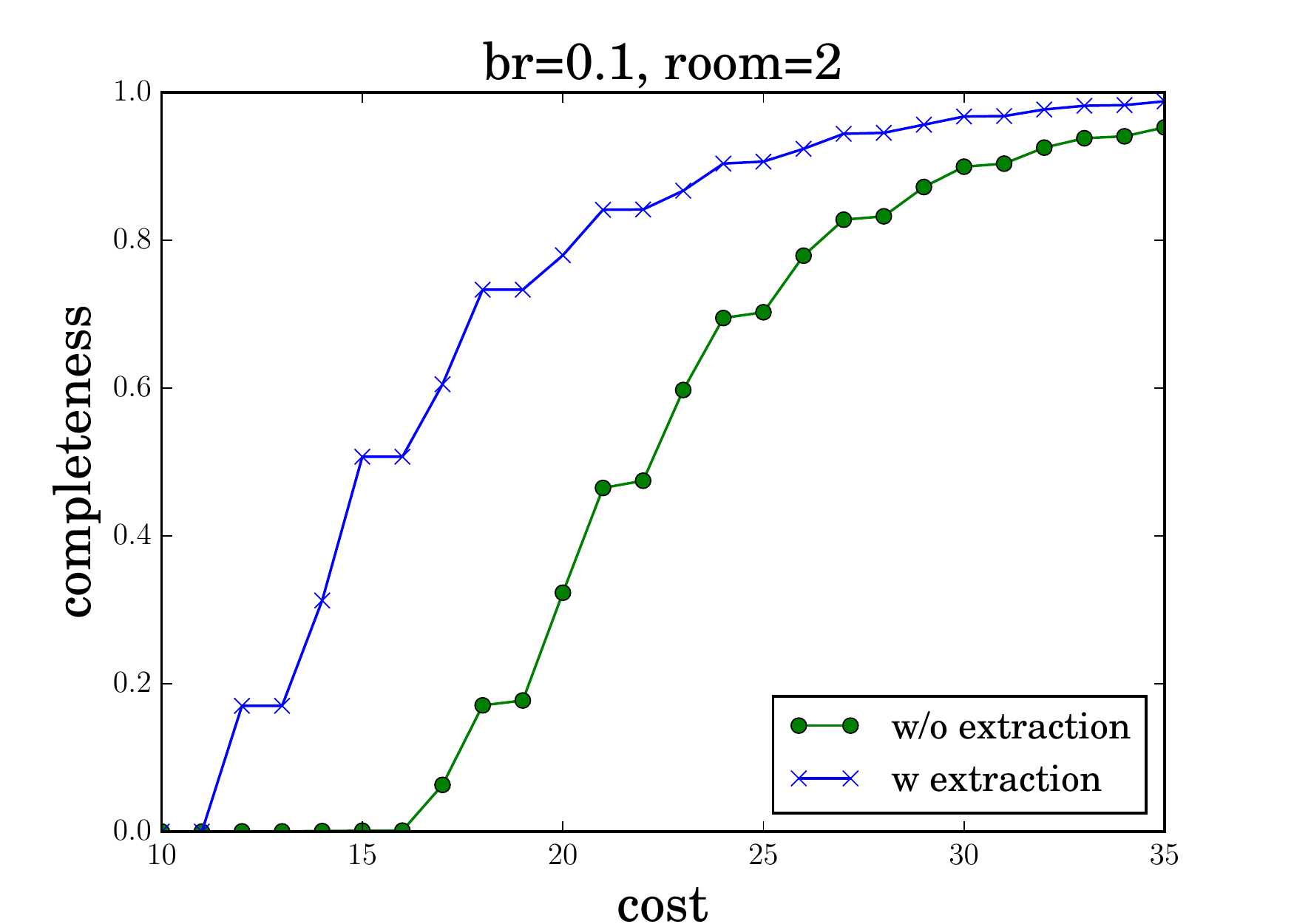}\hspace*{-1em}
    \includegraphics[width=0.26\textwidth]{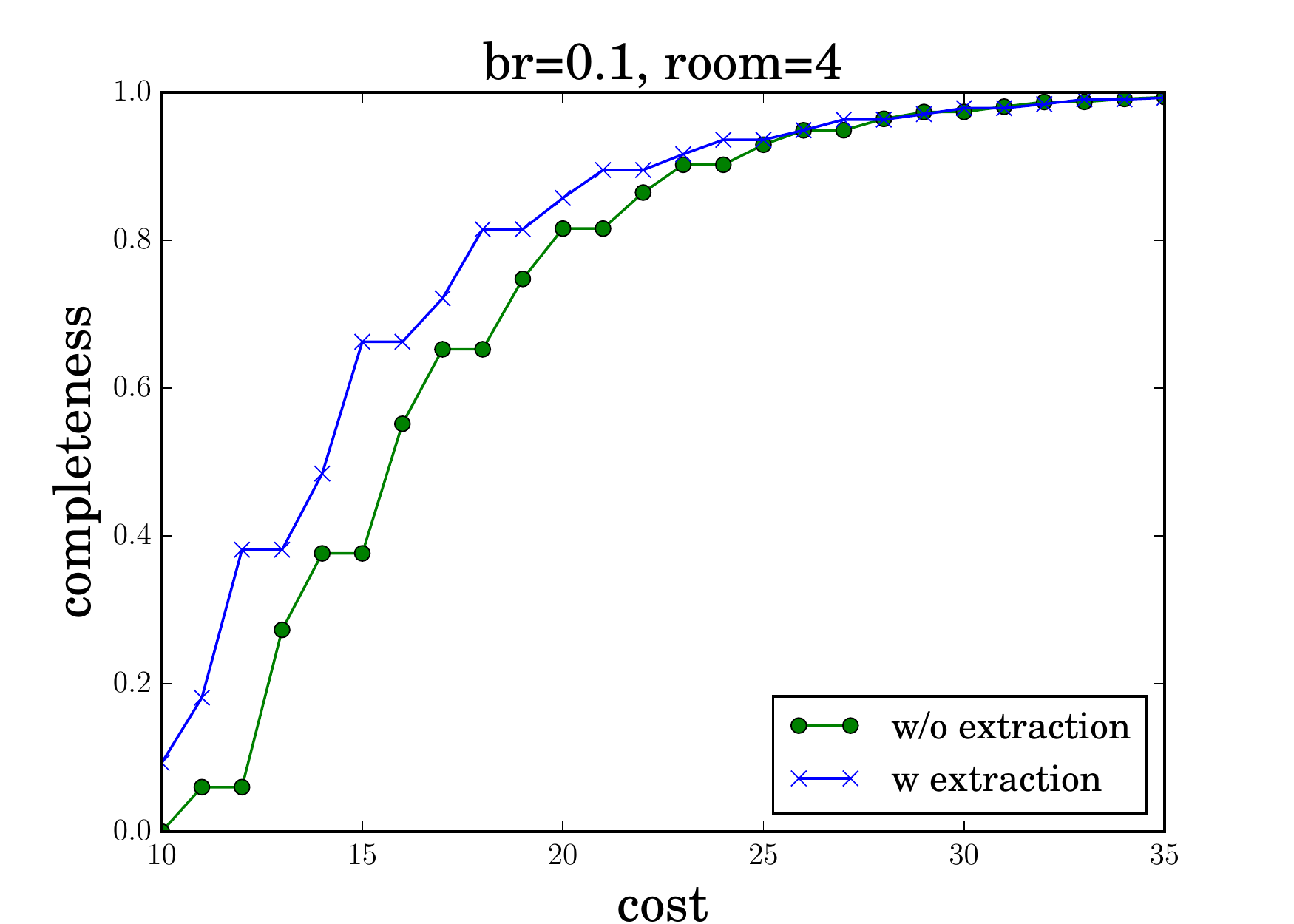}\\
    \hspace*{-.5em}
    \includegraphics[width=0.26\textwidth]{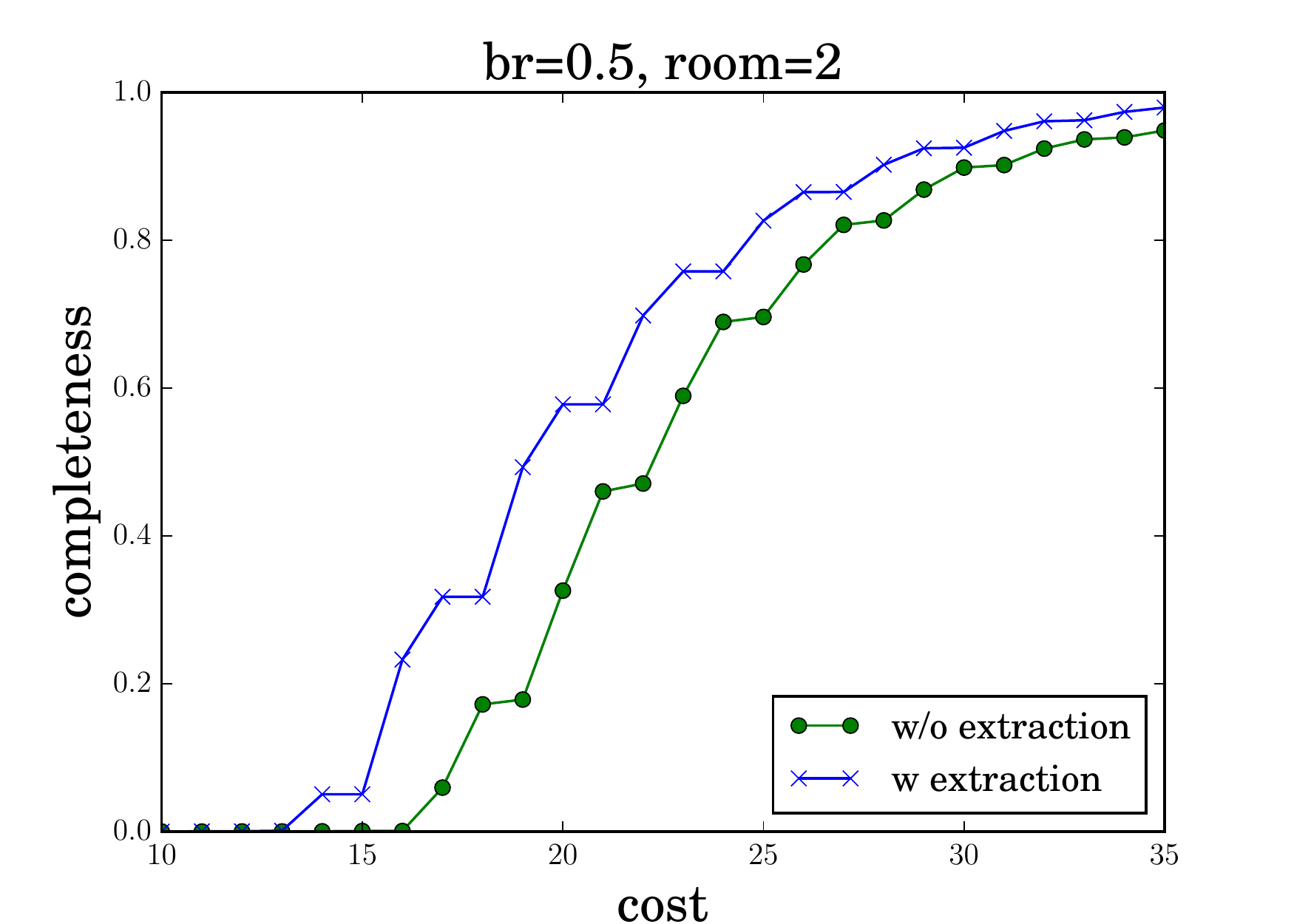}\hspace*{-1em}
    \includegraphics[width=0.26\textwidth]{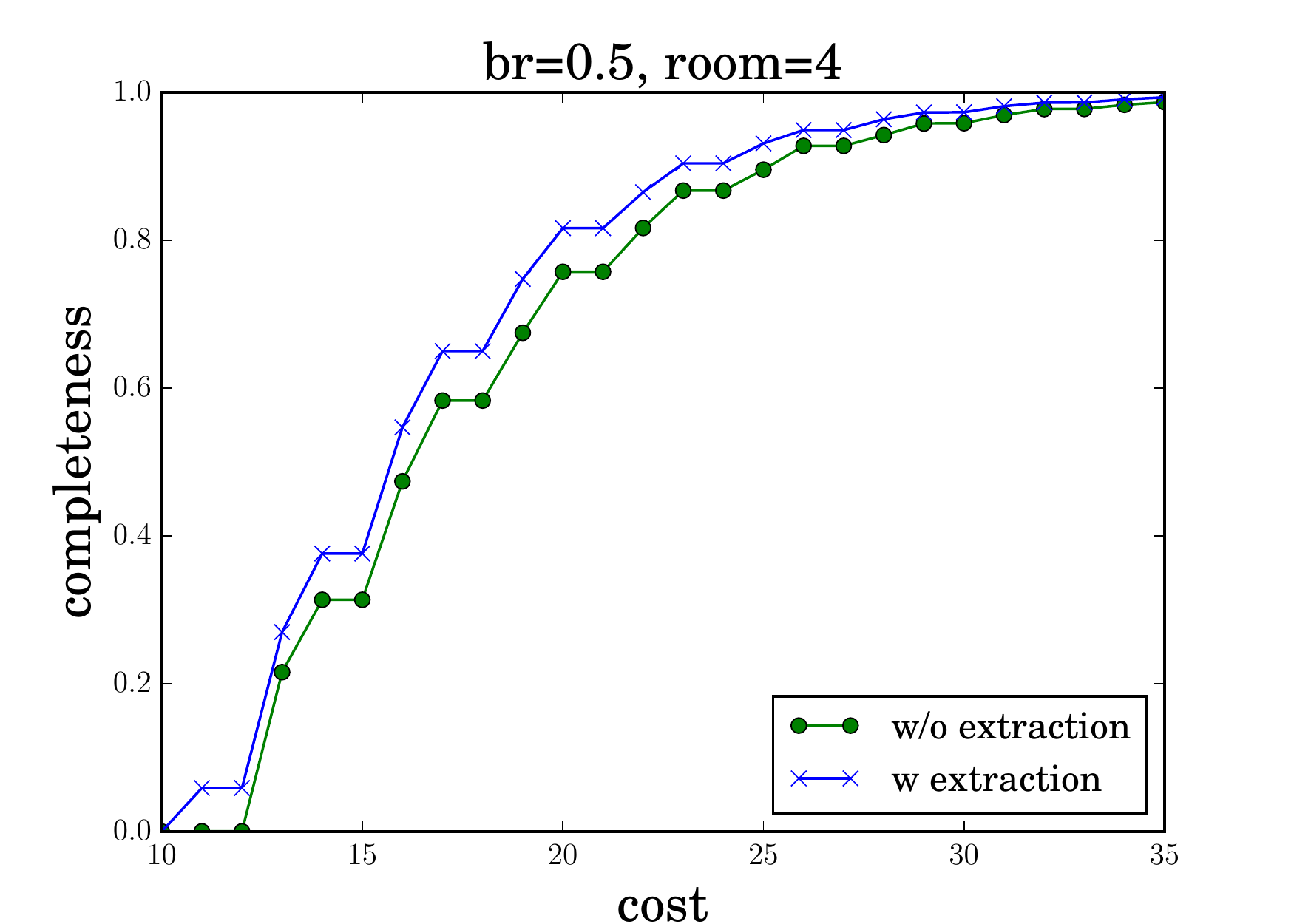}

  \end{center}
  \vspace{-1em}
  \caption{CDF plots of dialog completions. The requests are deliveries to \c{room2} (Left) and \c{room4} (right). }
  \label{fig:completeness}
  \vspace{-.5em}
\end{figure}

\section{Additional Experiments}
\label{sec:experiment}

To better analyze robot dialog behaviors, we generate cumulative distribution function (CDF) plots showing the percentage of dialog completions (y-axis) given different QA costs (x-axis). Figure~\ref{fig:completeness} shows the results when the request are deliveries to \c{room2} (left) and \c{room4} (right), under different \emph{blocking rates} of $br=0.1$ (top) and $br=0.5$ (bottom). 
Comparing the two curves in each subfigure, we find our KRR-RL framework reduces the QA cost in dialogs, which is consistent to Table~1 reported in the main paper.
Comparing the left (or right) two subfigures, we find a lower dialog completion rate given a higher $br$.
For instance, when the QA cost is 20, the percentage of dialog completion is reduced from 0.8 to 0.6 when \emph{br} grows from 0.1 to 0.5.
This indicates that our dialog manager becomes more conservative given higher blocking rates, which meets our expectation.
Comparing the top (or bottom) two subfigures, we find a smaller QA cost is needed, when the request is a delivery to \c{room4} (c.f., \c{room2}). This observation makes sense, because \c{room4} is closer to the shop (see Figure~3 in the main paper) and deliveries to \c{room4} is easier.

\end{document}